%% file: meta-learning_continual.tex
\documentclass{article}
\usepackage[pdftex]{graphics}
\usepackage[numbers]{natbib}

\usepackage[preprint]{nips_2018}

\usepackage[utf8]{inputenc} 
\usepackage[T1]{fontenc}    
\usepackage{hyperref}       
\usepackage{url}            
\usepackage{booktabs}       
\usepackage{amsfonts}       
\usepackage{nicefrac}       
\usepackage{microtype}      
\usepackage{algorithm}
\usepackage{algpseudocode}
\usepackage{amsmath}
\usepackage{graphicx}
\usepackage{subcaption}
\usepackage{courier}

\title{Meta Continual Learning}

\author{
  Risto Vuorio, Dong-Yeon Cho, Daejoong Kim, and Jiwon Kim \\
  SK T-Brain\\
  \texttt{\{vuoristo, dycho24, djkim, jk\}@sktbrain.com} \\
}

\begin{document}

\maketitle

\begin{abstract}
Using neural networks in practical settings would benefit from the ability of the networks to learn new tasks throughout their lifetimes without forgetting the previous tasks. This ability is limited in the current deep neural networks by a problem called catastrophic forgetting, where training on new tasks tends to severely degrade performance on previous tasks. One way to lessen the impact of the forgetting problem is to constrain parameters that are important to previous tasks to stay close to the optimal parameters. Recently, multiple competitive approaches for computing the importance of the parameters with respect to the previous tasks have been presented. In this paper, we propose a learning to optimize algorithm for mitigating catastrophic forgetting. Instead of trying to formulate a new constraint function ourselves, we propose to train another neural network to predict parameter update steps that respect the importance of parameters to the previous tasks.  In the proposed meta-training scheme, the update predictor is trained to minimize loss on a combination of current and past tasks. We show experimentally that the proposed approach works in the continual learning setting.
\end{abstract}

\input{Sec1_introduction.tex}
\input{Sec2_related_work.tex}
\input{Sec3_our_approach.tex}
\input{Sec4_experiments.tex}
\section{Discussion}
In this paper, we proposed a learning to optimize approach to continual learning. To the best of our knowledge, this is the first time that mitigating catastrophic forgetting has been explored in the learning to optimize context. Through the proposed approach, the update step prediction model successfully learned how to guide the optimization of neural network parameters for continual learning. We showed the feasibility of our meta-learning algorithm for continual learning by applying it to sequential image classification problems.

Currently our meta-training and continual training both use pixel permutations of the MNIST dataset. The different permutations still have the same number of samples from each class and the input distributions are closely related, even though they are not the same. It is likely that a well tuned update predictor would be able to exploit similarities in the meta-training and meta-testing datasets. We acknowledge this limitation in our experimental setting and leave exploration of the generalization ability of our approach for future work.

In our experiments we limit our learning setting to a maximum of three sequential tasks. In general continual learning, we would naturally be interested in learning on much longer tasks sequences. Our current update step predictor is designed to only consider the previous task in the sequence. Information about further back tasks could be incorporated in the optimization process by for example chaining multiple update predictors, each with their own previous task to consider or summarizing previous task information with an RNN. Handling longer sequences would be a promising avenue of future work. Note that our meta continual learning setting would trivially generalize to longer task sequences through sampling meta-training samples from further back tasks.

Many recent advances in machine learning have come from the move from hand designed features to learned ones. The optimizers we use to learn the features have themselves been subject to parameterization and learning. Learning to optimize, however, has proven challenging. In \cite{LV17} some central challenges of learning to optimize are presented and methods for improving the learned optimizer performance are presented. In this work, we worked in the intersection of two challenging problems: continual learning and learning to optimize. We found learning to optimize challenging in its own right, but we note that the advances presented in \cite{LV17} are largely orthogonal to our approach, and therefore could be applied to improve our update step predictor.



\bibliographystyle{abbrv}
\bibliography{meta-learning_continual}

\end{document}

%% file: Sec1_introduction.tex
\section{Introduction}
While we do not know exactly how humans learn throughout their lifetime, it is obvious that humans can continually accumulate information obtained by study or experience and efficiently develop new skills by using acquired knowledge. That is, distinguishing traits of human learning are retainment of knowledge from the past learning and fast adaptation to new environments based on the previous experience.

Many artificial neural network models have taken their inspiration from biological neural networks, however, they struggle to adeptly deal with solving multiple tasks sequentially. The primary source of this weakness has long been known as catastrophic forgetting or interference \cite{McCloskey1989,Ratcliff1990,French1999} where learning a new task may alter some important connection weights for old tasks and, in turn, make networks lose their ability to do previous tasks. Many attempts have been made to alleviate this intricate problem resorting to (i) increase in network capacity \cite{Rusu2016} that handles new tasks without affecting learned networks, (ii) extra-memory \cite{Lopez-Paz2017} or generative models \cite{Shin2017} that sustainedly provide the data for past tasks, and (iii) consideration of a regularization term \cite{Kirkpatrick2017,Zenke2017,Aljundi2017} that restrains drastic changes of current model.

Most of the highest performing algorithms in continual learning belong to the third category. At a high-level, the regularization terms employed by these approaches work by limiting the per-parameter updates to reduce the negative impact of learning a new task on the previously learned tasks. While these approaches often achieve high performance on continual learning tasks, they are limited by the accuracy of the assumptions made by the algorithm designers. For example, approximating the sensitivity of the performance on previous tasks to changes in each individual parameter is a challenging problem and improvements in the third category have come from improving this approximation \cite{Zenke2017,Aljundi2017}. In order to design a truly general continual learning algorithm, it seems unlikely that hand-crafted approximations will be enough. To take a step into the direction of more general continual learning algorithms, inspired by the recent successes in meta-learning and learning to learn research \cite{Andrychowicz2016,Finn2017}, we explore automatically learning a learning rule for continual learning.
%
%

The purpose of this work is to show the feasibility of learning to learn without forgetting. Meta-learning for continual learning has not yet been directly explored in the literature. In \cite{Meier2018,Al-Shedivat2018}, only fast adaptation for similar incoming tasks has been explored without concern about forgetting issue. We propose a meta-learning approach to continual learning where a model for adjusting per-parameter update step is trained to mitigate forgetting of past tasks. The aim of this update step prediction model is to assign small updates for parameters that should have low flexibility to maintain performance on the previous tasks and large updates for parameters that can be changed freely for the current task.

This paper is organized as follows. In Section 2, we summarize related work. In Section 3, we describe the proposed meta-learning framework for continual learning with detailed learning algorithm to train the update step prediction model. In Section 4, experimental results are given. Finally, we conclude and offer some directions for future research in Section 5.

%% file: Sec2_related_work.tex
\section{Related work}
One of the simplest ways to prevent from catastrophic forgetting of learned knowledge and achieve fast adaptation to new tasks in continual learning is to consider the expansion of network structure \cite{Rusu2016,Yoon2018} such as adding neurons, layers or networks whenever a new task is given. The main drawback of this approach is that it requires the boundless increase in network capacity as the number of tasks continuously grows even though the network capacity can be expanded only when necessary as in \cite{Yoon2018}.

Many continual learning approaches \cite{Rebuffi2017,Lopez-Paz2017} employed a certain form of memory to store data from previous tasks. Rebuffi \textit{et al.} \cite{Rebuffi2017} trained their model with the augmented training set which consists of new data and stored examples by means of distillation to prevent catastrophic forgetting. In \cite{Lopez-Paz2017}, regarding the losses of the previous data in the memory as inequality constraints, the model parameter was updated for the current task through quadratic programming. This allows not only to alleviate forgetting of the past tasks but also to beneficially transfer the knowledges from the current task to previous ones.
Instead of explicit memory, a deep generative model to mimic past data is trained in the generative adversarial networks (GANs) framework \cite{Shin2017}. In \cite{Triki2017}, informative features rather than the data of the previous tasks was captured by an autoencoder. This framework was able to ensure the preservation of important features when facing a new task. Kemker and Kanan \cite{Kemker2018} proposed FearNet architecture that consists of three brain-inspired neural networks for recent storage, long-term storage, and determining which one should be used for a particular example, respectively. Sprechmann \textit{et al.} \cite{Sprechmann2018} employed an embedding network as well as output network to generate keys for training data. For a given query, keys and their output values stored in a memory were retrieved in a context-based way at a test time. Then, they were used to locally adapt the parameters of the output network.

Researchers also tried to alleviate catastrophic forgetting by estimating how important each weight is for the previous tasks. After the pioneering work by \cite{Kirkpatrick2017}, where a regularizer to protect important parameters was computed as some coefficient times the squared difference of the current weight values and the last weight values from training on the previous task, some variations for computing the coefficient were proposed. Originally, the coefficient was an approximation of the Fisher information \cite{Kirkpatrick2017}, which is computed from the gradients of the loss function. In \cite{Zenke2017}, the coefficient was computed based on the contribution to the total reduction in loss and changes in parameter space over the entire trajectory of training. In \cite{Aljundi2017}, the coefficient was computed from the gradients of the trained model output on unlabeled as well as training data points. From the Bayesian inference point of view, Nguyen \textit{et al.} \cite{Nguyen2018} elaborated these approaches by combining online variational inference with Monte Carlo variational inference for neural networks.

Considering multiple models rather than a single model for sequentially given tasks, Lee \textit{et al.} \cite{Lee2017} proposed the method called incremental moment matching (IMM) to combine the models which had been individually trained on each task so that the final model had the ability to carry out multiple tasks. They tried to make the combined model balance the old and new tasks by approximating a mode of the mixture of two Gaussian posteriors with the assistance of a drop-out based transfer technique.

Using conceptors originally proposed in \cite{Jaeger2014}, He and Jaeger \cite{He2018} adjusted the gradients given by backpropagation so that learning a new task would minimally interfere with previously learned tasks. Whereas this variant of the backpropagation was formulated by human experts, some learning-to-learn or meta-learning approaches successfully designed optimizers for update rules in an automatic way \cite{Andrychowicz2016,Li2017,LV17}. However, meta-learning for continual learning has not yet been directly explored in the literature. A problem setting similar to continual learning, but without concern for the forgetting problem has been explored in two meta-learning papers. In \cite{Meier2018}, an alternative to constraining the weight updates was explored by learning a memory for learning rates to facilitate fast adaptation on similar tasks encountered subsequently. Fast adaptation through meta-learning was also explored in \cite{Al-Shedivat2018}, where an RL agent faces a sequence of opponents with increasing difficulty.

%% file: Sec3_our_approach.tex
\section{Our approach}
One of the most typical deep learning scenarios is supervised learning, where we want to find parameters $\theta$ of the input-output mapping function $f$ for the given task $\mathcal{T}$. During training, these parameters are updated iteratively, $\theta^{new} = \theta^{old} + \Delta\theta_t$. Gradient based optimization algorithms search the effective update step $\Delta\theta$ by considering the gradient $g=\nabla_{\theta}\mathcal{L}(f_{\theta}(\mathcal{T}))$ of a loss function $\mathcal{L}$ that measures distance between the output of the function $f$ and that of given task. In SGD, for example, $\Delta\theta$ is set to $-\alpha g$, where $\alpha$ is the learning rate which can be regarded as a shared hyperparameter across all parameters in the model. As in \cite{Andrychowicz2016}, the optimizers learned in an automatic way also used this gradient as an input to predict the optimal update step. Although the learned optimizers achieve faster convergence in a limited test setting than hand-crafted optimizers, there is little chance to avoid catastrophic forgetting without thoughtful design of a method to carefully control the updates.

Instead of relying on human experts for developing superior intelligent systems, meta-learning algorithms try to make AI systems learn how to learn. Under the general meta-learning umbrella, many different algorithms have been proposed in various areas of deep learning. These algorithms include, for example, good initialization of parameters for fast adaptation to new data \cite{Finn2017,Nichol2018} and improved optimization of gradient descent techniques \cite{Andrychowicz2016}. To the best of our knowledge, however, there is no study of a meta-learning algorithm to alleviate catastrophic forgetting in continual learning.

\begin{figure}
\includegraphics[width=12cm]{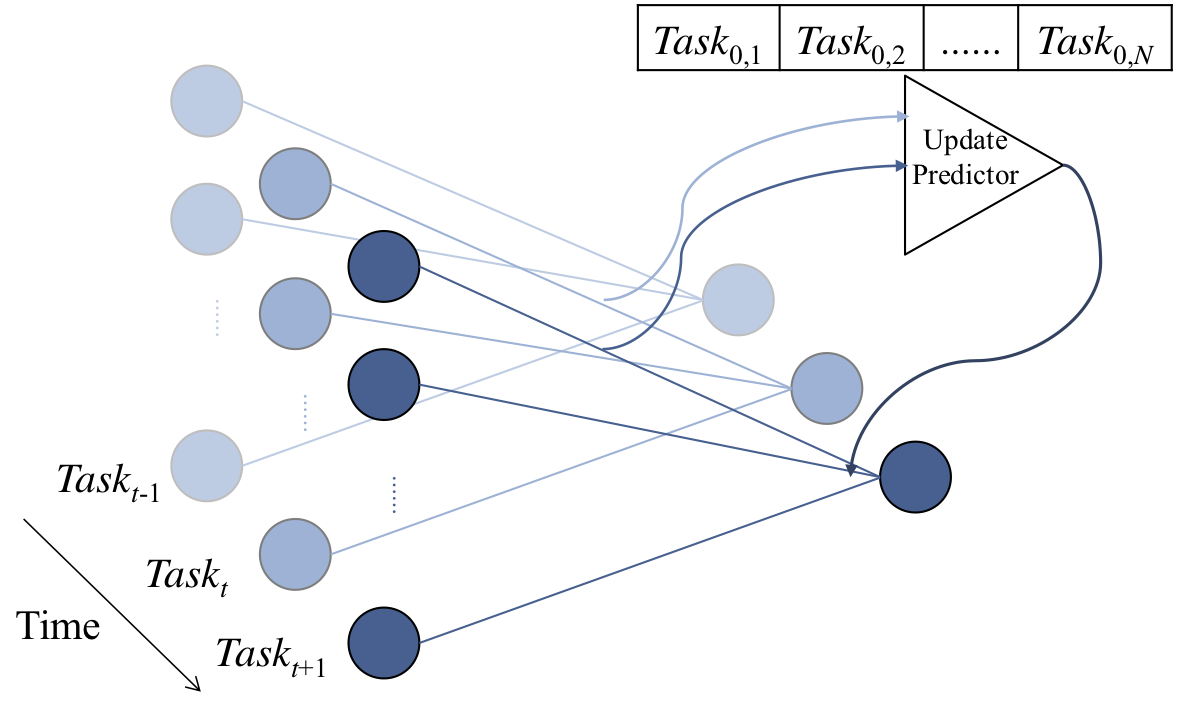}
\centering
\caption{During meta-learning step, the proposed update prediction model was trained on the given task $\mathcal{T}_0$ which consists of some subtasks. Using this model, we successfully learned a deep neural network over the sequentially given tasks without catastrophic forgetting.}
\label{fig1}
\end{figure}

In this paper, we present a meta-learning approach to continual learning, where we consider an update step prediction model $h$ represented by a multilayer perceptron or an RNN (Figure \ref{fig1}). The model is trained to limit the updates for parameters of the mapping function $f$ that are important for performance on the previous tasks and allow large updates for parameters used to learn the current task. Our prediction model is similar to the learnable optimizer in \cite{Andrychowicz2016} with the distinction that our optimizer is specialized into continual learning through its inputs and the meta-training setting.

In the ideal case, the update step predictor would be able to consider all of the mapping function parameters simultaneously. However, such approach is not feasible as even very simple deep neural networks often have tens of thousands of parameters. Instead, and again similarly to \cite{Andrychowicz2016}, our update step predictor operates on the mapping function parameters element-wise. 

When a new task is given to the mapping function $f$ for continual learning, the function's parameters $\theta$ are updated using the outputs of the update step prediction model as follows:
\begin{equation}
\begin{split}
g^*_{j-1} &= \nabla_{\theta^*}\mathcal{L}(f_{\theta^*}(\mathcal{T}_{j-1})),\\
g_{j} &= \nabla_{\theta}\mathcal{L}(f_{\theta}(\mathcal{T}_{j})) ,\\
\Delta\theta &= h_{\phi}(g^*_{j-1},g_{j},\mathcal{I}) ,\\
\theta' &= \theta - \eta\Delta\theta ,\\
\end{split}
\end{equation}
where $j$ indexes the tasks in the continual learning sequence, $\theta^*$ are the (local) optimal parameters for the previous task, $g^*_{j-1}$ are the average squared gradients of the previous task computed at $\theta^*$, $h_{\phi}$ is the update step prediction function parameterized by $\phi$, the scaling factor $\eta$ is a hyperparameter and $\mathcal{I}$ denotes the inputs to predict update step per parameter. In order to enable the update step predictor to output suitable updates considering the current and the previous tasks, it should be able to compute the importance of each mapping function parameter for the previous task compared to the other parameters. To make this possible, the update step predictor is fed the previous model parameters, the average squared gradients of the previous task and the current model parameters as inputs in addition to the current gradient.



The update step predictor parameters $\phi$ can be trained using a conventional optimization algorithm such as SGD or Adam. The loss for the update predictor is computed over the combination of previous and current tasks. Computing this loss requires backpropagation through the last step of the update prediction model parameter as follows:
\begin{equation}
\phi \leftarrow \texttt{Adam} (\nabla_{\phi}\mathcal{L}(f_{\theta}(\mathcal{T}_{j-1}\cup \mathcal{T}_{j}))),
\end{equation}
where we optimize over the union of the current and previous meta-training tasks. The motivation for this formulation is that, after the prediction model parameter update, we want to (i) achieve high performance on the current task (ii) retain high performance on the previously encountered tasks. Minimizing the loss on both previous and current tasks makes the update step predictor to assign small outputs to parameters that are central to the previous tasks and large outputs to parameters that are flexible to the current task. In this work, we focus on training of our update step prediction model as well as demonstrating how the prediction model can be used to keep important parameters for previous tasks unchanged in sequential classification problems. However, many interesting variations of our method are possible. For example, we can have the model prepare for fast transfer learning on future tasks considering the next task $\mathcal{T}_{0,j+1}$ during the meta continual learning.

The outline of the meta continual learning is presented in Algorithm \ref{code1}. For training our update step prediction model, we consider an independent meta-training dataset $\mathcal{T}_0$, which contains subtasks similar to the target continual learning tasks. Assuming that these subtasks are given sequentially, $h_\phi$ is iteratively optimized over two consecutive subtasks as explained above. During this meta-training, we expect that $h_\phi$ gains the ability to control per-parameter update step for preventing catastrophic forgetting. Note that in our algorithm, we train the mapping function on the first task of each task sequence using a standard optimizer such as Adam because in the beginning of the sequence there is nothing to preserve.

After the completion of meta-training, we start continual learning for the given tasks by using the trained $h_{\phi^*}$ function (Algorithm \ref{code2}). Similarly as in our meta-training setting, we first train a deep neural network $f_\theta$ over $\mathcal{T}_1$ by Adam optimizer. Then, $f_\theta$ is continually trained over the following tasks not by using a conventional optimizer for deep neural networks but by using the learned update prediction model $h_{\phi^*}$ from Algorithm \ref{code1}.

\begin{algorithm}
\caption{Meta continual learning for training $h_\phi$}\label{code1}
\begin{algorithmic} 
\Procedure{META-CONTINUAL-LEARNING}{$f_{\theta}, h_{\phi}, \mathcal{T}_0$}
\ForAll{$\mathcal{T}_{0,j>1}$ in $\mathcal{T}_0$}
\For{Epochs 1,2,3, ...}
\State $\theta_{0,j-1}^{*} \leftarrow $ train $f_{\theta}$ for one epoch on $\mathcal{T}_{0,j-1}$ using \texttt{Adam}
\State $\theta \leftarrow \theta_{0,j-1}^{*}$
\State $g_{j-1} \leftarrow \nabla_{\theta}\mathcal{L}(f_{\theta}(\mathcal{T}_{0,j-1}))$
\For{Epochs 1,2,3, ...}
\State $g \leftarrow \nabla_{\theta}\mathcal{L}(f_{\theta}(\mathcal{T}_{0,j}))$
\State $\theta \leftarrow \theta - \eta h_{\phi}(g_{j-1},g,\theta_{0,j-1}^{*},\mathcal{I})$
\State $\phi \leftarrow$ \texttt{Adam} $(\nabla_{\phi}\mathcal{L}(f_{\theta}(\mathcal{T}_{0,j-1}\cup \mathcal{T}_{0,j})))$
\EndFor
\EndFor
\EndFor
\EndProcedure
\end{algorithmic}
\end{algorithm}

\begin{algorithm}
\caption{Continual learning using the trained $h_{\phi^*}$}\label{code2}
\begin{algorithmic} 
\Procedure{Continual-Learning}{$f_{\theta}, h_{\phi^*}, \{\mathcal{T}_1,\mathcal{T}_2,\cdots\}$}
\ForAll{$\mathcal{T}_i$}
\If{$i=1$} 
\State $\theta_1^* \leftarrow$ Train $f_{\theta_1}$ on $\mathcal{T}_1$ using \texttt{Adam}
\Else
\State $\theta \leftarrow \theta_{i-1}^{*}$
\State $g_{i-1} \leftarrow \nabla_{\theta}\mathcal{L}(f_{\theta}(\mathcal{T}_{i-1}))$
\For{Epochs 1,2,3, ...}
\State $g \leftarrow \nabla_{\theta}\mathcal{L}(f_{\theta}(\mathcal{T}_i))$
\State $\theta \leftarrow \theta - \eta h_{\phi^*}(g_{i-1},g,\theta_{i-1}^{*},\mathcal{I})$
\EndFor
\State $\theta_i^* \leftarrow \theta$
\EndIf
\EndFor
\EndProcedure
\end{algorithmic}
\end{algorithm}

Following the recent learning to optimize approaches, a natural extension of our idea would be to use recurrent neural networks to parameterize the update step predictor. The meta-optimization process for the recurrent update step predictor would be the same as for the feed-forward predictor for the most part, except backpropagation through time would be used to optimize the recurrent model. We conducted preliminary experiments using an LSTM based update step predictor, but the results were inconclusive and are omitted from the paper.

%% file: Sec4_experiments.tex
\section{Experiments}
In recent research \cite{He2018,Lopez-Paz2017,Shin2017}, continual learning has been most commonly applied to classification problems such as variations of images from the MNIST handwritten digit database. Using a variation of the MNIST dataset, we carried out experiments to validate the proposed algorithm as a feasible approach for mitigating catastrophic forgetting.

In continual learning a model is trained over a sequence of tasks. In meta-learning, a distribution over tasks is used to meta-train a model, which is expected to generalize to other tasks from the same distribution. Due to the nature of both continual learning and meta-learning, we need access to a distribution of tasks. To generate a suitable distribution over tasks, we follow a problem setting previously considered in continual learning \cite{NIPS2013_5059,goodfellow2013empirical}, where the tasks are pixel permutations of the MNIST dataset. In this setting, each task, both for the meta-training and continual learning, is generated by shuffling the pixels of the original dataset by a fixed random permutation. The setting gives us access to numerous datasets of equal size and difficulty, where a model trained on single task is expected to not be able solve the other tasks without finetuning.

In our experimental setting we use a fully-connected neural network as the classification model. The is trained on the sequence of pixel shuffled tasks. The classification network has two hidden layers of 800 ReLU units. The update step predictor is another fully connected neural network with two layers of ten neurons. 

For each problem we ran meta-training (Algorithm \ref{code1}) with ten random initializations and picked the best update prediction model as measured by the validation accuracy. We then evaluated the trained models on ten newly sampled problem instances through Algorithm \ref{code2}.

\subsection{Disjoint MNIST}
\label{subsec:DisjointMNIST}
\begin{figure}
\centering
\begin{subfigure}{.5\textwidth}
  \centering
  \includegraphics[width=.9\linewidth]{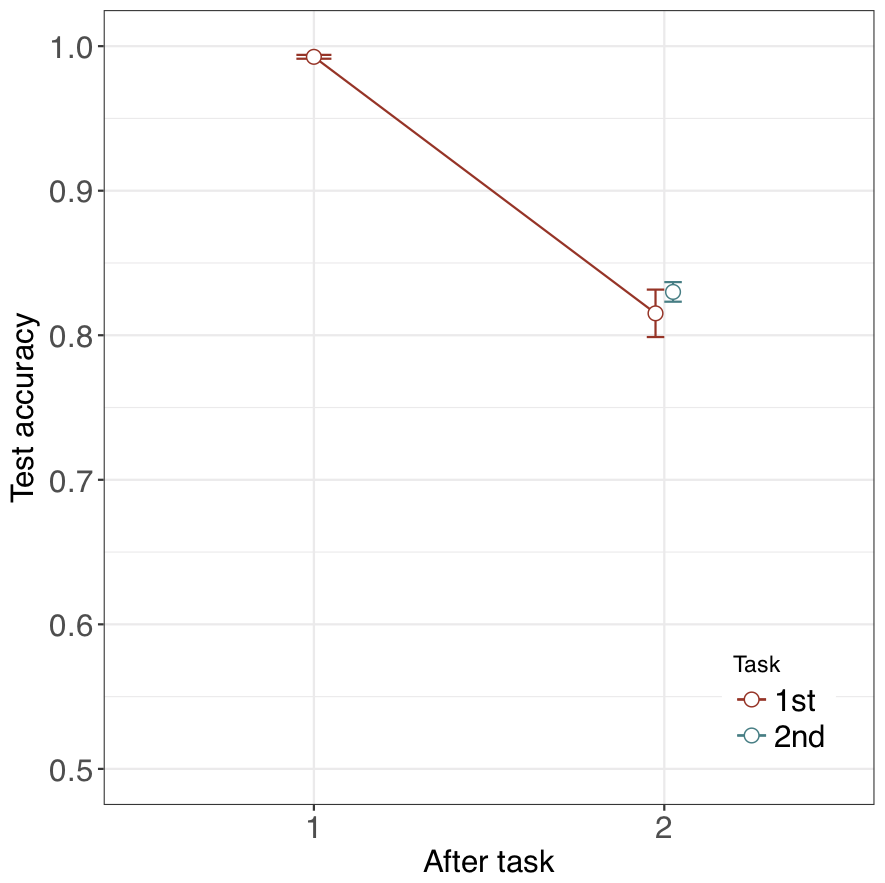}
  \label{fig:sub1}
\end{subfigure}%
\begin{subfigure}{.5\textwidth}
  \centering
  \includegraphics[width=.9\linewidth]{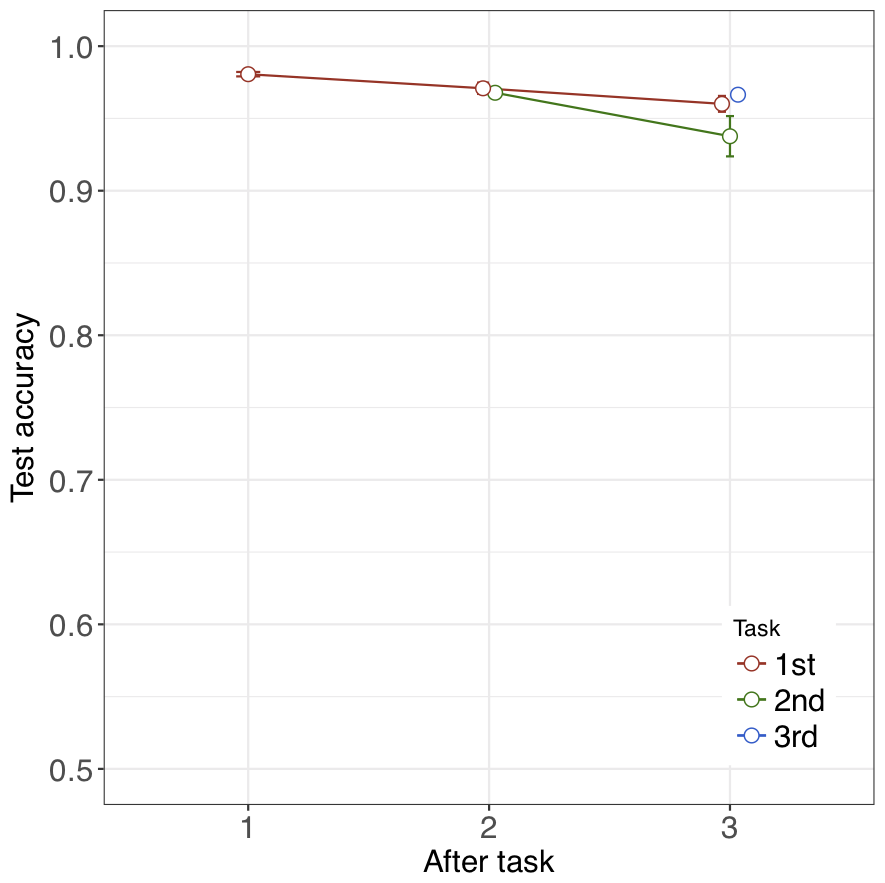}
  \label{fig:sub2}
\end{subfigure}
\caption{(Left) Results for disjoint MNIST (Right) three tasks from shuffled MNIST with our MLP predictor}
\label{fig2}
\end{figure}

In this section, we show the feasibility of the proposed algorithm by looking into the behavior of the learned update step prediction model on two sequential tasks. We use the Disjoint MNIST problem setting from \cite{NIPS2013_5059}. In the Disjoint MNIST problem setting, MNIST classes are split into two disjoint sets, one consisting of images from $\{0, 1, 2, 3, 4\}$ and the other from $\{5, 6, 7, 8, 9\}$. Similarly to \cite{Lee2017} we consider the problem of ten class joint classification. For the meta-training phase, we use another MNIST permutation, which has been split into two tasks similarly to the target task. Results from the experiment are presented in Table \ref{tab:results1}. The results demonstrate that our model is capable of significantly reducing the catastrophic forgetting compared to the baseline of training the prediction model on one task and then the other using SGD. This indicates that the trained update step prediction model has an ability to deal with current task while preserving knowledge about the previous task. We also confirmed this by checking the test accuracy of each task after finishing continual learning (Figure \ref{fig2}).

\begin{figure}
\includegraphics[width=14cm]{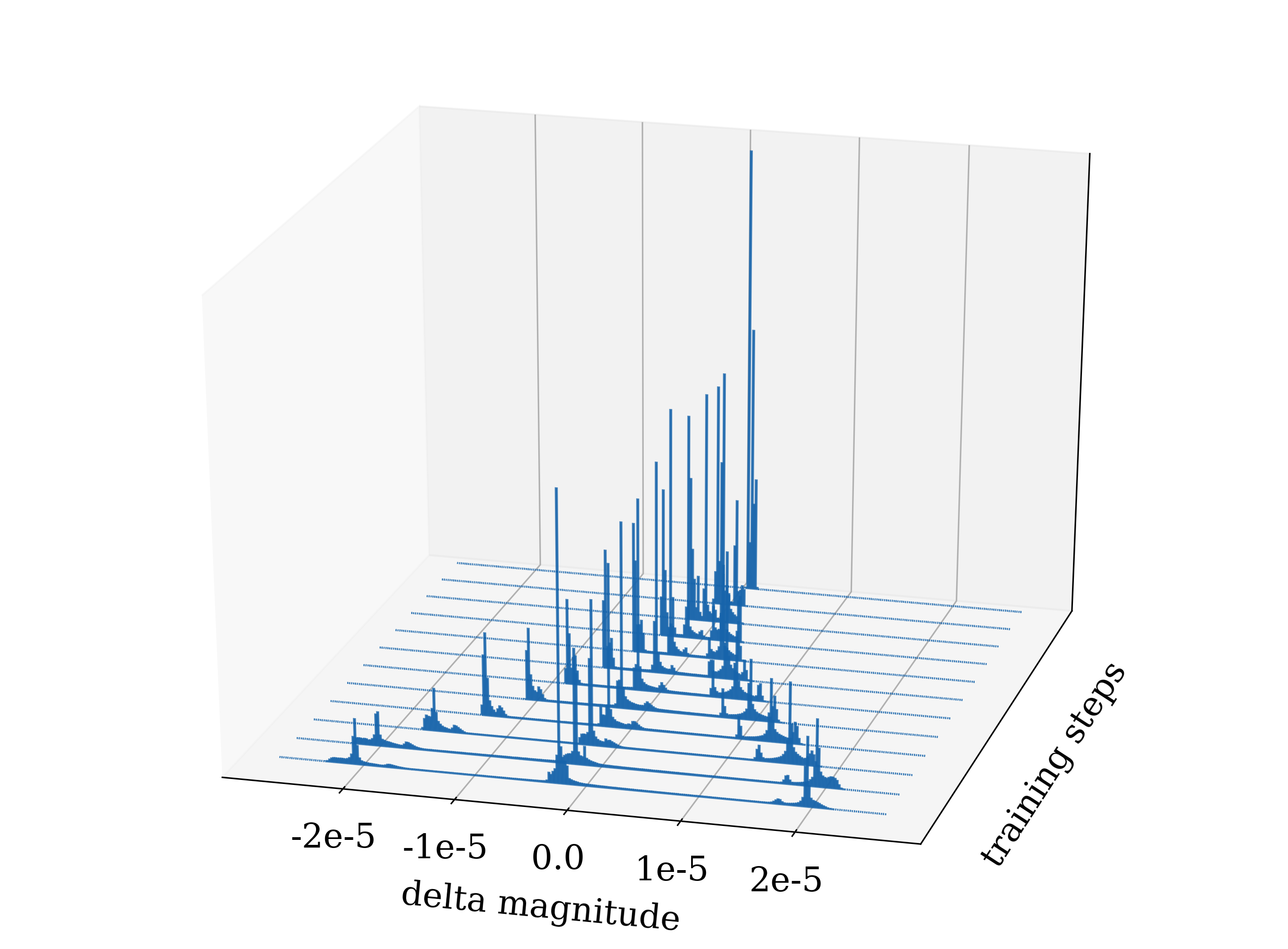}
\centering
\caption{Evolution of the update predictor output (multiplied by the scaling factor $\eta$) of our update step prediction model}
\label{fig3}
\end{figure}

\begin{table}
\begin{center}
\begin{tabular}{l|ll}
\toprule
Method & disjoint MNIST & shuffled MNIST\\
\midrule
SGD (untuned)& 47.72 $\pm$ 0.11 &  89.15$\pm$2.34 \\
ours (MLP) & 82.3 $\pm$ 0.92& 95.5 $\pm$ 0.58\\
\bottomrule
\end{tabular}
\end{center}
\caption{Averaged test accuracy (\%) $\pm$ standard deviation over 10 runs. Numbers for SGD are excerpts from \cite{Lee2017}.}
\label{tab:results1}
\end{table}

\subsection{Multi-task setting}
\label{subsec:MultiMNIST}
We conduct experiments on sequences of three shuffled MNIST tasks \cite{goodfellow2013empirical,NIPS2013_5059}. In this setting, the continual learning tasks are three pixel permutations of MNIST. Similarly to the Disjoint MNIST setting, we train the update predictor using another set of shuffled datasets. In this experimental setting, the state-of-the-art models achieve performance close to the joint training over all tasks \cite{Lee2017}. While not being competitive with the other models in terms of performance, our model demonstrates that it has learned to preserve the knowledge of the previous task as shown in Table \ref{tab:results2}.

To analyze the characteristics of the update step prediction model, we collected the output values of the model at different training stages. As shown in Figure \ref{fig3}, all output values were very closed to zero immediately after starting the meta-training. However, we had a clear observation of tri-modal distributions as the meta-training iterations went by. One possible elucidation on this change is as follows. At early stages of training, our update prediction model had little knowledge of the importance of parameters. The lack of knowledge makes the prediction model have very small output values. While training goes on, our model acquires the knowledge so it is able to supply the appropriate values for adjusting the parameters which are not crucial to the success of previous tasks.
\begin{table}
\begin{center}
\begin{tabular}{l|ll}
\toprule
Method & disjoint MNIST & shuffled MNIST\\
\midrule
SGD (tuned)& 71.32 $\pm$ 1.54 &  $\sim$95.5 \\
EWC & 52.72 $\pm$ 1.36 &  $\sim$98.2 \\
IMM (best) & 94.12 $\pm$ 0.27 & 98.3 $\pm$ 0.08 \\
\midrule
ours (MLP) & 82.3 $\pm$ 0.92& 95.5 $\pm$ 0.58\\
\bottomrule
\end{tabular}
\end{center}
\caption{Averaged test accuracy (\%) $\pm$ standard deviation over 10 runs. Numbers for other methods are excerpts from \cite{Lee2017}.}
\label{tab:results2}
\end{table}

%% file: meta-learning_continual.bbl
\begin{thebibliography}{10}

\bibitem{Al-Shedivat2018}
M.~Al-Shedivat, T.~Bansal, Y.~Burda, I.~Sutskever, I.~Mordatch, and P.~Abbeel.
\newblock Continuous adaptation via meta-learning in nonstationary and
  competitive environments.
\newblock In {\em International Conference on Learning Representations}, 2018.

\bibitem{Nichol2018}
J.~S. Alex~Nichol, Joshua~Achiam.
\newblock On first-order meta-learning algorithms.
\newblock {\em CoRR}, abs/1803.02999, 2018.

\bibitem{Aljundi2017}
R.~Aljundi, F.~Babiloni, M.~Elhoseiny, M.~Rohrbach, and T.~Tuytelaars.
\newblock Memory aware synapses: Learning what (not) to forget.
\newblock {\em CoRR}, abs/1711.09601, 2017.

\bibitem{Andrychowicz2016}
M.~Andrychowicz, M.~Denil, S.~G$\grave{o}$mez, M.~W. Hoffman, D.~Pfau,
  T.~Schaul, and N.~de~Freitas.
\newblock Learning to learn by gradient descent by gradient descent.
\newblock In D.~D. Lee, M.~Sugiyama, U.~V. Luxburg, I.~Guyon, and R.~Garnett,
  editors, {\em Advances in Neural Information Processing Systems 29}, pages
  3981--3989. Curran Associates, Inc., 2016.

\bibitem{Finn2017}
C.~Finn, P.~Abbeel, and S.~Levine.
\newblock Model-agnostic meta-learning for fast adaptation of deep networks.
\newblock In D.~Precup and Y.~W. Teh, editors, {\em Proceedings of the 34th
  International Conference on Machine Learning}, pages 1126--1135,
  International Convention Centre, Sydney, Australia, 06--11 Aug 2017. PMLR.

\bibitem{French1999}
R.~M. French.
\newblock Catastrophic forgetting in connectionist networks.
\newblock {\em Trends in cognitive sciences}, 3(4):128--135, 1999.

\bibitem{goodfellow2013empirical}
I.~J. Goodfellow, M.~Mirza, D.~Xiao, A.~Courville, and Y.~Bengio.
\newblock An empirical investigation of catastrophic forgetting in
  gradient-based neural networks.
\newblock In {\em International Conference on Learning Representations}, 2014.

\bibitem{He2018}
X.~He and H.~Jaeger.
\newblock Overcoming catastrophic interference using conceptor-aided
  backpropagation.
\newblock In {\em International Conference on Learning Representations}, 2018.

\bibitem{Jaeger2014}
H.~Jaeger.
\newblock Controlling recurrent neural networks by conceptors.
\newblock {\em CoRR}, abs/1403.3369, 2014.

\bibitem{Kemker2018}
R.~Kemker and C.~Kanan.
\newblock Fearnet: Brain-inspired model for incremental learning.
\newblock In {\em International Conference on Learning Representations}, 2018.

\bibitem{Kirkpatrick2017}
J.~Kirkpatrick, R.~Pascanu, N.~Rabinowitz, J.~Veness, G.~Desjardins, A.~A.
  Rusu, K.~Milan, J.~Quan, T.~Ramalho, A.~Grabska-Barwinska, D.~Hassabis,
  C.~Clopath, D.~Kumaran, and R.~Hadsell.
\newblock Overcoming catastrophic forgetting in neural networks.
\newblock {\em Proceedings of the National Academy of Sciences},
  114(13):3521--3526, 2017.

\bibitem{Lee2017}
S.-W. Lee, J.-H. Kim, J.~Jun, J.-W. Ha, and B.-T. Zhang.
\newblock Overcoming catastrophic forgetting by incremental moment matching.
\newblock In I.~Guyon, U.~V. Luxburg, S.~Bengio, H.~Wallach, R.~Fergus,
  S.~Vishwanathan, and R.~Garnett, editors, {\em Advances in Neural Information
  Processing Systems 30}, pages 4652--4662. Curran Associates, Inc., 2017.

\bibitem{Li2017}
K.~Li and J.~Malik.
\newblock Learning to optimize.
\newblock In {\em International Conference on Learning Representations}, 2017.

\bibitem{Lopez-Paz2017}
D.~Lopez-Paz and M.~A. Ranzato.
\newblock Gradient episodic memory for continual learning.
\newblock In I.~Guyon, U.~V. Luxburg, S.~Bengio, H.~Wallach, R.~Fergus,
  S.~Vishwanathan, and R.~Garnett, editors, {\em Advances in Neural Information
  Processing Systems 30}, pages 6467--6476. Curran Associates, Inc., 2017.

\bibitem{LV17}
K.~Lv, S.~Jiang, and J.~Li.
\newblock Learning gradient descent: Better generalization and longer horizons.
\newblock In D.~Precup and Y.~W. Teh, editors, {\em Proceedings of the 34th
  International Conference on Machine Learning}, pages 2247--2255,
  International Convention Centre, Sydney, Australia, 06--11 Aug 2017. PMLR.

\bibitem{McCloskey1989}
M.~McCloskey and N.~J. Cohen.
\newblock Catastrophic interference in connectionist networks: The sequential
  learning problem.
\newblock {\em Psychology of Learning and Motivation}, 24:109--165, 1989.

\bibitem{Meier2018}
F.~Meier, D.~Kappler, and S.~Schaal.
\newblock Online learning of a memory for learning rates.
\newblock In {\em IEEE International Conference on Robotics and Automation},
  2018.

\bibitem{Nguyen2018}
C.~V. Nguyen, Y.~Li, T.~D. Bui, and R.~E. Turner.
\newblock Variational continual learning.
\newblock In {\em International Conference on Learning Representations}, 2018.

\bibitem{Ratcliff1990}
R.~Ratcliff.
\newblock Catastrophic interference in connectionist networks: The sequential
  learning problem.
\newblock {\em Psychological Review}, 97(2):285--308, 1990.

\bibitem{Rebuffi2017}
S.-A. Rebuffi, A.~Kolesnikov, G.~Sperl, and C.~H. Lampert.
\newblock i{C}a{RL}: Incremental classifier and representation learning.
\newblock In {\em Proceedings of IEEE Conference on Computer Vision and Pattern
  Recognition}, pages 5533--5542, Honolulu, HI, USA, 21--26 Jul 2017. IEEE.

\bibitem{Rusu2016}
A.~A. Rusu, N.~C. Rabinowitz, G.~Desjardins, H.~Soyer, J.~Kirkpatrick,
  K.~Kavukcuoglu, R.~Pascanu, and R.~Hadsell.
\newblock Progressive neural networks.
\newblock {\em CoRR}, abs/1606.04671, 2016.

\bibitem{Shin2017}
H.~Shin, J.~K. Lee, J.~Kim, and J.~Kim.
\newblock Continual learning with deep generative replay.
\newblock In I.~Guyon, U.~V. Luxburg, S.~Bengio, H.~Wallach, R.~Fergus,
  S.~Vishwanathan, and R.~Garnett, editors, {\em Advances in Neural Information
  Processing Systems 30}, pages 2990--2999. Curran Associates, Inc., 2017.

\bibitem{Sprechmann2018}
P.~Sprechmann, S.~Jayakumar, J.~Rae, A.~Pritzel, A.~P. Badia, B.~Uria,
  O.~Vinyals, D.~Hassabis, R.~Pascanu, and C.~Blundell.
\newblock Memory-based parameter adaptation.
\newblock In {\em International Conference on Learning Representations}, 2018.

\bibitem{NIPS2013_5059}
R.~K. Srivastava, J.~Masci, S.~Kazerounian, F.~Gomez, and J.~Schmidhuber.
\newblock Compete to compute.
\newblock In C.~J.~C. Burges, L.~Bottou, M.~Welling, Z.~Ghahramani, and K.~Q.
  Weinberger, editors, {\em Advances in Neural Information Processing Systems
  26}, pages 2310--2318. Curran Associates, Inc., 2013.

\bibitem{Triki2017}
A.~R. Triki, R.~Aljundi, M.~B. Blaschko, and T.~Tuytelaars.
\newblock Encoder based lifelong learning.
\newblock {\em CoRR}, abs/1704.01920, 2017.

\bibitem{Yoon2018}
J.~Yoon, E.~Yang, J.~Lee, and S.~J. Hwang.
\newblock Lifelong learning with dynamically expandable networks.
\newblock In {\em International Conference on Learning Representations}, 2018.

\bibitem{Zenke2017}
F.~Zenke, B.~Poole, and S.~Ganguli.
\newblock Continual learning through synaptic intelligence.
\newblock In D.~Precup and Y.~W. Teh, editors, {\em Proceedings of the 34th
  International Conference on Machine Learning}, pages 3987--3995,
  International Convention Centre, Sydney, Australia, 06--11 Aug 2017. PMLR.

\end{thebibliography}
